\documentclass[sigconf]{acmart}
\AtBeginDocument{%
  }

\acmConference[ICAIL'26]{21th International Conference on Artificial Intelligence and Law}{June 8--16, 2026}{Singapore}

\usepackage{multirow}
\usepackage{scalerel}
\usepackage{algorithm}
\usepackage{algpseudocode}
\usepackage{tikz}
\usepackage{pgfplots}
\usetikzlibrary{svg.path}
\usepackage{arydshln}
\pgfplotsset{compat=1.18}
\usepgfplotslibrary{statistics}

\definecolor{orcidlogocol}{HTML}{A6CE39}
\usepackage[table]{xcolor}

\usepackage{xcolor}

\tikzset{
  orcidlogo/.pic={
    \fill[orcidlogocol] svg{M256,128c0,70.7-57.3,128-128,128C57.3,256,0,198.7,0,128C0,57.3,57.3,0,128,0C198.7,0,256,57.3,256,128z};
    \fill[white] svg{M86.3,186.2H70.9V79.1h15.4v48.4V186.2z}
                 svg{M108.9,79.1h41.6c39.6,0,57,28.3,57,53.6c0,27.5-21.5,53.6-56.8,53.6h-41.8V79.1z M124.3,172.4h24.5c34.9,0,42.9-26.5,42.9-39.7c0-21.5-13.7-39.7-43.7-39.7h-23.7V172.4z}
                 svg{M88.7,56.8c0,5.5-4.5,10.1-10.1,10.1c-5.6,0-10.1-4.6-10.1-10.1c0-5.6,4.5-10.1,10.1-10.1C84.2,46.7,88.7,51.3,88.7,56.8z};
  }
}

\newcommand\orcidicon[1]{\href{https://orcid.org/#1}{\mbox{\scalerel*{
\begin{tikzpicture}[yscale=-1,transform shape]
\pic{orcidlogo};
\end{tikzpicture}
}{|}}}}

\usepackage{hyperref} 

\begin{document}

\title{Legal2LogicICL: Improving Generalization in Transforming Legal Cases to Logical Formulas via Diverse Few-Shot Learning}

\author{Jieying Xue}
\authornote{Both authors contributed equally to this paper.}
\orcid{0009-0000-8070-6609}
\affiliation{%
  \institution{Center for Juris-Informatics, ROIS-DS, \\Tokyo, Japan}
  \city{}
  \country{}
}
\email{xuejieying@jaist.ac.jp}

\author{Phuong Minh Nguyen}
\authornotemark[1]
\orcid{0000-0002-3752-8699}
\affiliation{%
  \institution{Japan Advanced Institute of Science and Technology, Ishikawa, Japan}
  \city{}
  \country{}
}
\email{phuongnm@jaist.ac.jp}

\author{Ha Thanh Nguyen}
\orcid{0000-0003-2794-7010}
\affiliation{%
  \institution{Center for Juris-Informatics, ROIS-DS}
  \city{Tokyo}
  \country{Japan}
}
\email{nguyenhathanh@nii.ac.jp}

\author{May Myo Zin}
\orcid{0000-0003-1315-7704}
\affiliation{%
  \institution{Center for Juris-Informatics, ROIS-DS}
  \city{Tokyo}
  \country{Japan}
}
\email{maymyozin@nii.ac.jp}

\author{Ken Satoh}
\orcid{0000-0002-9309-4602}
\affiliation{%
  \institution{Center for Juris-Informatics, ROIS-DS, \\Tokyo, Japan}
  \city{}
  \country{}
}
\email{ksatoh@nii.ac.jp}


\begin{abstract}
This work aims to improve the generalization of logic-based legal reasoning systems by integrating recent advances in Natural Language Processing (NLP) with legal-domain adaptive few-shot learning techniques using Large Language Models (LLMs). Existing logic-based legal reasoning pipelines typically rely on fine-tuned models to map natural-language legal cases into logical formulas before forwarding them to a symbolic reasoner. However, such approaches are heavily constrained by the scarcity of high-quality annotated training data.
To address this limitation, we propose a novel LLM-based legal reasoning framework that enables effective in-context learning through retrieval-augmented generation (RAG). 
Specifically, we introduce \textit{Legal2LogicICL}, a few-shot retrieval framework that balances diversity and similarity of exemplars at both the \textit{latent semantic representation} level and the \textit{legal text structure} level. In addition, our method explicitly accounts for legal structure by mitigating entity-induced retrieval bias in legal texts, where lengthy and highly specific entity mentions often dominate semantic representations and obscure legally meaningful reasoning patterns. 
Our \textit{Legal2LogicICL} constructs informative and robust few-shot demonstrations, leading to accurate and stable logical rule generation without requiring additional training.
In addition, we construct a new dataset, named \textbf{Legal2Proleg}, which is annotated with alignments between legal cases and PROLEG logical formulas to support the evaluation of legal semantic parsing. Experimental results on both open-source and proprietary LLMs 
demonstrate that our approach significantly improves accuracy, stability, and generalization in transforming natural-language legal case descriptions into logical representations, highlighting its effectiveness for interpretable and reliable legal reasoning\footnote{Our code is available at \href{https://github.com/yingjie7/Legal2LogicICL}{https://github.com/yingjie7/Legal2LogicICL}.}. 

\end{abstract}

%




\maketitle
\section{Introduction} 
Legal reasoning is an essential task of the legal domain, in which legal judgments are derived by interpreting statutory provisions together with case facts. However, the scale and complexity of legal systems, along with the ambiguity of natural language case descriptions, make the formalization of legal reasoning highly challenging. In particular, accurately transforming unstructured legal texts into structured, machine-interpretable logical representations remains a critical problem in legal AI, as it directly affects the reliability of automated legal reasoning systems.
To support interpretable legal reasoning, the PROLEG framework was proposed \citep{satoh2009translating, Satoh2023}, providing an expressive formalism that enables legal professionals to understand reasoning processes and outcomes. However, PROLEG-based systems require inputs in the form of formal logical formulas, which creates a usability barrier for practitioners without expertise in logical modeling.
To mitigate this issue, prior work has introduced multi-stage legal reasoning pipelines that translate natural language case descriptions into PROLEG fact formulas before inference \citep{nguyen2022multi}. In such pipelines, the semantic parsing stage is crucial, as errors in this step directly propagate to downstream reasoning.
Existing approaches for this task can be broadly categorized into pattern-based methods \citep{navas2018contractframes}, neural machine translation models \citep{10.1007/978-3-031-21743-2_35,nguyen2022multi}, and NER-based systems \citep{zin2023improving}. While these methods have achieved partial success, they suffer from limited generalization, high annotation costs, or brittleness to surface-level variations, particularly when handling complex legal structures and multi-entity relations.

Recent advances in LLMs offer a promising alternative, as LLMs exhibit strong reasoning and generation capabilities without task-specific training.
In this work, we propose \textit{Legal2LogicICL}, a novel LLM-based legal semantic parsing framework that translates natural language legal case descriptions into PROLEG logical fact formulas through in-context learning, without requiring additional supervised fine-tuning. Our framework leverages the strong reasoning and generation capabilities of LLMs, while introducing structured guidance through carefully constructed in-context demonstrations. 
Specifically, we adopt a RAG paradigm \citep{NEURIPS2020_6b493230} to retrieve relevant legal cases as in-context exemplars. In the In-Context Learning (ICL) setting, where model parameters remain fixed, the quality of retrieved exemplars plays a critical role in determining reasoning performance \citep{liu-etal-2022-makes,MINHNGUYEN2025114256}. 
Prior studies have shown that effective few-shot sets should be both relevant to the target instance and sufficiently diverse to avoid overfitting to superficial patterns \citep{min-etal-2022-rethinking, zhang2022active, rubin-etal-2022-learning, shin-etal-2021-constrained,levy-etal-2023-diverse}. Overly homogeneous exemplars may cause the model to overfit to surface-level similarities, thereby limiting robustness and generalization.

Motivated by these observations, we propose a legal-oriented diversity-enhanced few-shot selection strategy, which explicitly controls exemplar diversity through two complementary mechanisms. 
First, we introduce a diversity control parameter $\lambda$ to regulate the trade-off between semantic similarity and exemplar diversity during retrieval.
Second, we address a fundamental limitation of conventional semantic similarity-based retrieval methods in the legal domain. Legal texts frequently contain long and highly specific entity mentions (e.g., party names, assets, contract identifiers, and descriptions of legal actions), which dominate semantic representations and bias retrieval toward surface-level entity overlap rather than meaningful legal structures.
To mitigate this issue, we introduce an entity-agnostic, template-level similarity strategy. Instead of computing similarity over raw legal cases, we abstract legal texts into templates by removing concrete entity instantiations while preserving their underlying legal and logical structure.  This enables the retrieval of structurally aligned exemplars even when surface entities differ.  By combining semantically similar case-level exemplars with structurally similar template-level exemplars, our \textit{Legal2LogicICL} constructs a diversity-aware hybrid few-shot set that balances contextual relevance and structural diversity.
This prompting strategy provides flexible structural constraints while preserving the expressive power of LLMs in language understanding and generation, enabling robust domain adaptation without model fine-tuning. Experimental results demonstrate that \textit{Legal2LogicICL} effectively mitigates entity-induced retrieval bias, improves generalization across diverse legal cases, and yields more accurate and stable generation of PROLEG-style logical rules. 
Overall, this work offers a principled and practical solution for converting natural language legal facts into structured logical representations and lays a solid foundation for future research on logic-based legal reasoning and decision-making.

\section{Related Work}
\subsection{Transforming Legal Text into Logical Forms}
Transforming natural language legal text into formal logical representations is essential for legal reasoning systems. Early approaches primarily rely on template-based methods that extract entities or facts using predefined rules \citep{bajwa2011sbvr, mccarty2007deep, lagos2010event, gaur2014translating}. While effective under constrained settings, such methods suffer from limited scalability and poor generalization to complex legal scenarios.

To improve applicability, \citet{nguyen2022multi} introduced a multi-stage translation framework that combines end-to-end translation models with an additional error-correction module. Although pre-trained models enhance flexibility, they remain vulnerable to overfitting and often fail to produce precise fine-grained details (e.g., temporal expressions), which are critical for logical reasoning. \citet{zin2023improving} further introduce a NER-based pipeline that enforces logical well-formedness via rule composition, but its reliance on surface forms makes it brittle under paraphrasing.

\subsection{Diversity in Few-shot In-Context Learning}
Few-shot in-context learning (ICL) has shown strong performance in legal reasoning due to its scalability and independence from task-specific training. Prior work highlights the importance of demonstration selection, particularly semantic similarity between examples and queries \citep{pasupat-etal-2021-controllable, liu-etal-2022-makes, rubin-etal-2022-learning}.
Beyond similarity, recent studies emphasize the role of diversity in improving generalization \citep{gupta-etal-2022-structurally, levy-etal-2023-diverse, MINHNGUYEN2025114256, cohen2024diversity}. Diverse demonstrations help cover a broader structural space, which is especially important in compositional tasks where models must generate structured outputs (e.g., logical forms). In such settings, limited structural coverage can hinder generalization due to missing symbolic patterns.


Taken together, prior studies indicate that diverse few-shot demonstrations provide more effective guidance for in-context learning by expanding the hypothesis space exposed to the model, thereby enabling stronger and more robust generalization.
Motivated by these findings, we propose a diversity-aware few-shot selection framework tailored to the legal domain. While existing approaches emphasize semantic similarity or structural diversity in general NLP settings, legal reasoning poses additional challenges due to its reliance on specialized terminology, symbolic constraints, and heterogeneous reasoning patterns. Our method explores domain-specific strategies for increasing demonstration diversity, with the goal of exposing in-context learners to a broader and more representative set of legal reasoning structures, thereby enhancing generalization in few-shot settings.

\section{Task Definition and Notations}
\label{sec:problem_definition}

We study the task of \textit{legal semantic parsing}, which aims to transform natural-language legal case descriptions into structured logical representations (e.g., a set of PROLEG fact formulas) that are executable within a legal reasoning system.

\paragraph{Legal Cases and Entities.}
Let $l \in \mathcal{L}$ denote a natural-language legal case (example in Table~\ref{tab:dataset_example}).
A legal case typically involves a set of concrete \textit{entities}, such as legal parties (e.g., borrower, lender), objects (e.g., assets), agreements, temporal expressions, and legally relevant events.
We denote the entity set associated with a legal case $l$ as
\begin{align}
\mathcal{E}(l) = \{ e_1, e_2, \ldots, e_{|\mathcal{E}(l)|} \},
\end{align}
where each entity $e_i$ corresponds to a specific real-world instantiation appearing in the case description.
\paragraph{Legal Templates.}
In the legal reasoning domain, different legal cases may share similar underlying legal structures while differing substantially in their concrete entities.
To capture such structural regularities, we define a \textit{legal template} as an entity-agnostic abstraction of a legal case.
Formally, given a legal case $l$, a template function  maps $l$ to a template $t = \mathrm{template}(l)$ by replacing each concrete entity mention in $l$ with its corresponding typed placeholder (or entity type)  (e.g., \texttt{\{Borrower\}}, \texttt{\{Lender\}}, \texttt{\{Agreement\}}).
\begin{align}
\mathrm{template}(\cdot): \mathcal{L} \rightarrow \mathcal{T}
\end{align}
The resulting template preserves the narrative structure and legally meaningful relations of the case while suppressing entity-specific surface information.

\paragraph{Fact Formula Generation.}
Given a legal template and a concrete entity set, a specific legal case can be constructed by filling the template placeholders with entities from $\mathcal{E}(l)$.
The goal of legal semantic parsing in this work is, given a legal case $l$, to identify the appropriate instantiations of PROLEG \textit{fact formulas} that satisfy the predefined legal rules.
We denote the target output as
\begin{align}
f = \{ f_1, f_2, \ldots, f_{|f|} \} \in \mathcal{F},
\end{align}
where each $f_i$ is an executable PROLEG fact corresponding to an entity-grounded legal predicate, as shown in the \textit{Facts} field of Table~\ref{tab:dataset_example}.

\paragraph{Task Formulation.}
Formally, the task can be defined as learning a transforming method ($\mathcal{M}$) which maps a natural-language legal case $l$ to its corresponding set of fact formulas $f$.
\begin{align}
\mathcal{M}: \mathcal{L} \rightarrow \mathcal{F},
\end{align}
In this work, the mapping $\mathcal{M}$ is induced through few-shot in-context learning using LLMs, without any task-specific parameter fine-tuning.
Instead, the model is guided by a small set of \textit{retrieved legal demonstrations} that provide both semantically relevant and structurally diverse reasoning patterns.

\begin{table}[htbp]
\caption{An Example Legal Case and Its Structured Representation in the Dataset \textbf{Legal2Proleg}}
\label{tab:dataset_example}
\vspace{0.5em}
\centering
\resizebox{\linewidth}{!}{
    \begin{tabular}{p{1.2\linewidth}}
    \toprule
     \textbf{Content Example } - (denotaion)\\
    \midrule
    \cellcolor{gray!15}\textbf{Legal Issue}  \\
    \textit{The borrower's use of the assets in a manner that harms the lender's interests or reputation has damaged the lender's rights and reputation.}
    \\
    \midrule
    \cellcolor{gray!15}\textbf{Legal Case} - ($l^{query}$) \\
    Medical supplies were given to the hospital by the health organization as part of a supply agreement, intended for patient treatment. Instead, the hospital redistributed the supplies to external clinics without consent. This resulted in a shortage of supplies during a critical time, affecting patient care. These actions were discovered on 2023/08/20. Does the health organization have grounds for legal action to protect their reputation? 
    \\
    \midrule
    
    \cellcolor{gray!15}\textbf{Entities Set (or Slot Holders)} - $(\mathcal{E}(l^{query}))$\\
    \{
    \texttt{\textbf{"Borrower"}: "The hospital"},\;
    \texttt{\textbf{"Object"}: "medical supplies"},\;
    \texttt{\textbf{"Lender"}: "the health organization"},\;
    \texttt{\textbf{"Agreement"}: "a supply agreement"},\;
    \texttt{\textbf{"Harm"}: "a shortage of supplies during a critical time, affecting patient care"},\;
    \texttt{\textbf{"T\_discovery"}: "2023/08/20"}
    \} 
    \\
    \midrule
    
    \cellcolor{gray!15}\textbf{Template} - ($\textrm{template} (l^{query})$) \\
    \textit{\{Object\}} was given to \textit{\{Borrower\}} by \textit{\{Lender\}} as part of \textit{\{Agreement\}}, meant for patient treatment. Instead, \textit{\{Borrower\}} redistributed supplies to external clinics without consent. This resulted in \textit{\{Harm\}}. These actions were discovered on \textit{\{T\_discovery\}}. Does \textit{\{Lender\}} have grounds for legal action to protect their reputation? 
    \\\midrule
    
    \cellcolor{gray!15}\textbf{Rules} - ($\mathcal{R}$)\\
    \texttt{\textbf{right\_to\_legal\_action(\_Lender, \_Borrower, \_Object)} <=} \newline
     \textcolor{white}{\quad} \texttt{harm\_to\_lender\_rights(\_Borrower, \_Lender, \_Object),} \newline
     \textcolor{white}{\quad} \texttt{discovery\_of\_harm(\_Lender, \_Object, \_T\_discovery).} \\[0.3em]
    
   \texttt{\textbf{harm\_to\_lender\_rights(\_Borrower, \_Lender, \_Object)} <=} \newline
     \textcolor{white}{\quad} \texttt{borrower(\_Borrower), lender(\_Lender),} \newline
     \textcolor{white}{\quad} \texttt{owned\_by(\_Object, \_Lender),} \newline
     \textcolor{white}{\quad} \texttt{borrowing\_agreement(\_Agreement),} \newline
    \textcolor{white}{\quad} \texttt{use\_under\_agreement(\_Borrower, \_Object, \_Agreement),} \newline
     \textcolor{white}{\quad} \texttt{unlawful\_use(\_Borrower, \_Object),} \newline
     \textcolor{white}{\quad} \texttt{reputational\_damage(\_Lender, \_Harm).} \\[0.3em]
    
     \texttt{\textbf{unlawful\_use(\_Borrower, \_Object)} <=} \newline
     \textcolor{white}{\quad} \texttt{use\_under\_agreement(\_Borrower, \_Object, \_Agreement),} \newline
     \textcolor{white}{\quad} \texttt{violation\_of\_agreement(\_Borrower, \_Agreement).} \\[0.3em]
    
     \texttt{\textbf{reputational\_damage(\_Lender, \_Harm)} <= harm\_fact(\_Harm, \_Lender).} \\[0.3em]
    
    \texttt{\textbf{discovery\_of\_harm(\_Lender, \_Object, \_T\_discovery)} <=} \newline
     \textcolor{white}{\quad} \texttt{discovery\_fact(\_Lender, \_Object, \_T\_discovery).} \\\midrule
    
    \cellcolor{gray!15}\textbf{Facts in Legal Case} - ($f^{query}$)\\
     \texttt{borrower("The hospital").} \newline
     \texttt{lender("the health organization").} \newline
    \texttt{owned\_by("medical supplies", "the health organization").} \newline
     \texttt{borrowing\_agreement("a supply agreement").} \newline
    \texttt{agreement\_fact("The hospital", "medical supplies", "a supply agreement").} \newline
    \texttt{violation\_of\_agreement("The hospital", "a supply agreement").} \newline
     \texttt{harm\_fact("a shortage of supplies during a critical time", "the health organization")} \newline
    \texttt{discovery\_fact("the health organization", "medical supplies", "2023/08/20")}  \vspace{0.5em} \\
    \texttt{\%} \textit{Legal Query Formula} 
    \\
    \texttt{right\_to\_legal\_action("the health organization", "The hospital", "medical supplies").} \\
    \bottomrule
    \end{tabular}
}
\end{table}

\section{Dataset Construction}
\label{sec:Legal2Proleg}
\paragraph{Motivation} To evaluate the generalization ability of methods for transforming natural language legal cases into PROLEG format logical formulas within the Proleg system, we construct a new dataset named \textbf{Legal2Proleg}. This dataset is built by adopting a data-augmentation pipeline for legal cases introduced in \citet{phuogdata}, which contains multiple contract types, including loan, lease, purchase, and copyright contracts. Table~\ref{tab:legal2logicicl_dataset} compares the characteristics of our dataset with those of previous datasets in the legal domain. Overall, the prior dataset \textit{LegalCaseNER} \citep{zin2023improving} is designed to evaluate the ability to detect object (entity) names in contracts. Although it contains the largest number of samples, it exhibits the lowest diversity (i.e., the smallest template\footnote{The template of each sample was constructed by substituting entity values with corresponding entity names.} vocabulary size) and is limited to purchase contracts only.
In contrast, \textit{Legal2ProlegV0}\footnote{The original authors did not assign a name to their experimental dataset; we denote it as \textit{Legal2ProlegV0} to properly credit their contribution.} \citep{phuogdata} and our extended version, \textit{Legal2Proleg}, demonstrate greater diversity in terms of template vocabulary size, the number of legal issues, and the variety of unique facts.
Notably, neither \textit{LegalCaseNER} nor \textit{Legal2ProlegV0} control for template overlap between the training and testing sets, which introduces a risk of overfitting when applying supervised fine-tuning (SFT) methods. For example, BERT-based models, which strongly encode bidirectional contextual information, may memorize template-specific contexts during training and thus perform poorly on previously unseen templates. This issue is further examined in our experimental section. 
The construction of \textit{Legal2Proleg} dataset proceeds through the following stages.
\begin{table}[t]
\centering
\caption{Statistics of the \textit{Legal2Proleg} dataset and others.}
\label{tab:legal2logicicl_dataset}
\resizebox{\linewidth}{!}{
    \begin{tabular}{l p{0.26\linewidth}  p{0.26\linewidth}   p{0.26\linewidth}}
    \toprule
     \textbf{Dataset}&\textbf{LegalCaseNER}&\textbf{Legal2ProlegV0} & \cellcolor{gray!15} \textbf{Legal2Proleg} \\
    \midrule
    \textbf{Source}   &\citet{zin2023improving} & \citet{phuogdata}  &    \cellcolor{gray!15} (this work) \\
    \textbf{\#Samples ($\times1000$)}   & $\approx$6.3  &  $\approx$5  &    \cellcolor{gray!15} $\approx$1  \\
    \textbf{\#Templates}   & 947 &  240  &   \cellcolor{gray!15} 1057 \\
    \textbf{\#Template Vocab.}   & 264 &  907  &   \cellcolor{gray!15} 2897 \\
    \textbf{\#Entity Types} &13 & 27 &  \cellcolor{gray!15} 45 \\
    \textbf{\#Legal Issues} & 4&    9 & \cellcolor{gray!15} 24 \\
    \textbf{\#Unique Facts} & 6&    31 & \cellcolor{gray!15} 114 \\
    \textbf{\#Contract Types} & Purchase& Purchase, Lease, Loan,  Copyright & \cellcolor{gray!15} Purchase, Lease, Loan,   Copyright \\
    \textbf{Disclosed templ.}   & Yes &  Yes  &  \cellcolor{gray!15} No \\
    \bottomrule
    \end{tabular}
}
\end{table}

\subsubsection{Annotation of PROLEG Rules}
\label{sec:logical_tree}
The annotator is first provided with a legal issue and an example case description (Table~\ref{tab:dataset_example}). At this stage, the annotator and analyzes the constraints specified in the contract (or legal document),   based on this analysis, the annotator manually construct the logical causal relationships between contractual constraints and the relevant case facts using the PROLEG logic language. These relationships are represented as legal rules ($\mathcal{R}$) or PROLEG logical trees, as illustrated in Table~\ref{tab:dataset_example}.
All annotations are constructed under expert supervision and undergo careful verification and correction to ensure both logical correctness and legal consistency.

\subsubsection{Annotation of Facts in Legal Cases}
\label{sec:fact_generation}
Given the legal rules and legal case obtained in the previous step, this stage focuses on annotating legally relevant facts in each legal case using PROLEG formulas ($f$).
An example illustrating the alignment between PROLEG formulas and the corresponding legal case information is shown in Table~\ref{tab:dataset_example}.
In addition, the important \textit{entity types} (or \textit{slot holders}) are identified and collected for subsequent template-based legal case data augmentation \citep{phuogdata}.

\subsubsection{Data Augmentation of Legal Template}
\label{sec:data_construction}
Given the provided legal issue, a case description, and predefined entity types, annotators construct realistic contractual scenarios covering diverse types of breaches, and then generate multiple corresponding logical templates.
Following the augmentation pipeline of \citet{phuogdata}, the content of templates and entity pairs can be generated with assistance from LLMs (e.g., gpt-5), but certainly need to be confirmed or rejected by annotators familiar with the legal issue.  This design ensures structural consistency at the template level while maintaining substantive diversity at the case level, closely mirroring real-world legal reasoning settings. Finally, for each legal issue, numerous templates and entity pairs are synthesized and aligned with the corresponding legal rules and fact formulas, which serve as gold-standard data for machine learning systems.

\section{Proposed Method}

\begin{figure*}
    \centering
       \includegraphics[width=1\textwidth, keepaspectratio, 
            trim={0.cm 0.2cm 0.5cm  0cm}, page=1, clip=true]{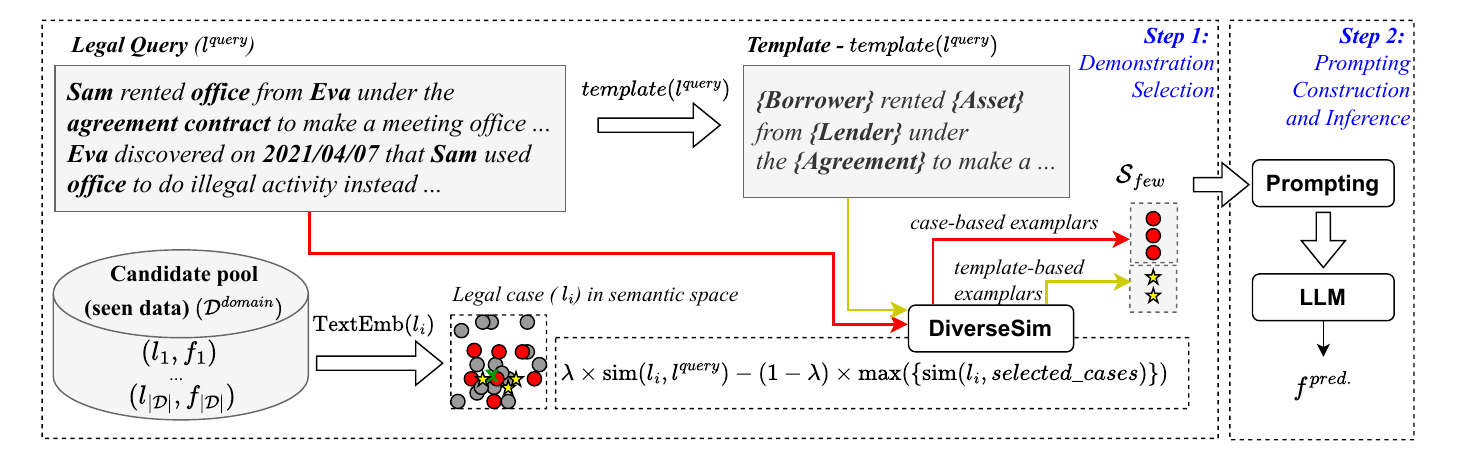}
    \caption{Overview of the proposed diversity-aware hybrid few-shot in-context learning framework.}
    \label{fig_overview}
\end{figure*}

As illustrated in Figure~\ref{fig_overview}, we propose \textit{Legal2LogicICL}, a diversity-oriented hybrid few-shot learning framework for legal translation from natural language into structured logical representations. Following the RAG \citep{NEURIPS2020_6b493230} and few-shot prompting paradigm \citep{NEURIPS2020_1457c0d6}, the framework constructs domain-aware demonstrations while incorporating diversity to guide LLMs toward structurally consistent and legally grounded reasoning, without task-specific fine-tuning.

We introduce a \textit{two-level diversity-similarity balancing mechanism}. At the latent semantic level, we propose \textit{DiverseSim}, a retrieval algorithm that selects demonstrations by regulating the trade-off between similarity and diversity via a tunable hyper-parameter over encoded vector representations. 
At the structural text level, we design a legal-domain-aware hybrid retrieval strategy. While conventional case-based retrieval is often dominated by surface-level entity overlap, we introduce an entity-agnostic template-based mechanism that emphasizes underlying legal structures. By combining semantically relevant exemplars with structurally aligned templates, our method enhances both diversity and robustness in in-context demonstrations.
We further decompose \textit{Legal2LogicICL} into two stages: (1) \textit{Demonstration Selection}, and (2) \textit{Prompting Construction and Inference}, which are detailed in the following subsections.

\subsection{Demonstration Selection \label{sec:demon_ranking}} 
    Motivated by prior work on diversity-aware retrieval for in-context learning \citep{levy-etal-2023-diverse, MINHNGUYEN2025114256}, we observe that two factors are particularly critical during demonstration retrieval: (1) semantic relevance to the query instance and (2) diversity among the selected demonstrations.
    Given a legal case query, the objective of this stage is to select demonstrations from the retrieval pool that are not only highly relevant to the query, but also mutually distinctive, so as to collectively cover a broad range of legal scenarios and reasoning patterns. This balance is especially important in the legal domain, where excessive similarity among demonstrations may bias the model toward narrow, entity-specific patterns and hinder generalization.
    Motivated by these considerations, we propose \textit{DiverseSim} (Algorithm~\ref{alg:selectcandidate}), a diversity-aware demonstration ranking algorithm tailored for the legal domain. \textit{DiverseSim} selects a set of few-shot examples that jointly maximize relevance to the query while promoting structural and contextual diversity, which are subsequently used to construct the in-context prompt for logical inference.
    
    \algrenewcommand\algorithmicrequire{\textbf{Input:}}
     \algrenewcommand\algorithmicensure{\textbf{Output:}}
    \begin{algorithm}
    \small
    \caption{\textit{DiverseSim}}\label{alg:cap} \label{alg:selectcandidate}
    \begin{algorithmic}[1]
    \Require a query legal case  $l^{query}$, a storage examples $\mathcal{D}^{domain} = \{ \langle   l_i, f_i\rangle\}_{0\leq i<|\mathcal{D}^{domain}|}$, number of candidates $k$, balancing weight of similarity and diversity scores $\lambda$. 
    \Ensure a set of selected demonstrations $\mathcal{S}$
    \State $\mathcal{S}  \gets  \{\}$   \Comment{\textcolor{blue}{init the set of selected demonstrations}}
    \State $\mathcal{D}  \gets  \mathcal{D}^{domain}$
    \State $ \mathbf{e}^{q} = \text{TextEmb}(l^{query}) $
    \State $\mathbf{e}^{d}_i = \text{TextEmb}(l_i)\quad \textrm{foreach} \quad  l_i \in \mathcal{D}$ 
    \State $\mathcal{B} = \operatorname{arg\,top\text{-}10}_{i < |\mathcal{D}|}\textrm{sim}(e_i^d, e^q)$  \Comment{\textcolor{blue}{most similar to query as a boundary}}
    \While{$|S| < k$}   \Comment{\textcolor{blue}{loop until enough demonstrations}}
        \State $\mathbf{e}^{s}_j = \text{TextEmb}(l_j)\quad \textrm{foreach} \quad  l_j \in \mathcal{S}$ 
        \State $rank_i = \lambda\times  \textrm{sim} (e_i^{d}, e^{q}) - (1-\lambda)\times \textrm{max}(\{\textrm{sim}(e_i^{d}, e_j^{s})\}_{j<|S|}) $ 
        \State $\mathcal{R} =\{rank_i\}_{i<|\mathcal{B}|}$    \Comment{\textcolor{blue}{ranking with redundancy penalty}}
         \State $\mathcal{S}\gets  \mathcal{S}  \cup \underset{ (l_i, f_i)}{\textrm{argmax}}(\mathcal{R})$ \Comment{\textcolor{blue}{pick best element based on ranking scores}}
         \State $\mathcal{B} \gets \mathcal{B} \setminus\mathcal{S}$ \Comment{\textcolor{blue}{avoid duplicated selections in next step}}
          
    \EndWhile
    \end{algorithmic}
\end{algorithm}

\subsubsection{Few-shot Selection via Query Content} 


Given a query legal case, we first retrieve the top-$n$ semantically similar legal cases from a domain-specific corpus to construct high-quality contextual demonstrations.
We hypothesize that legal cases exhibiting stronger semantic and structural similarity to the query are more likely to provide informative and reliable references for downstream logical reasoning. 
To operationalize this process, we construct a domain-specific corpus, denoted as $\mathcal{D}^{\mathrm{domain}} = \{ l_i \}_{i=1}^{N}
$, where each $l_i$ denotes a legal case in the training corpus.
Formally, given a query legal case $l^{\mathrm{query}}$, we retrieve the top-$n$ most semantically similar legal cases from $\mathcal{D}^{\mathrm{domain}}$. 
During retrieval, semantic similarity is computed using cosine similarity between vector representations of legal cases.
All candidate query-example pairs are ranked according to \textit{DiverseSim} algorithm, and the top-$n$ ranked cases are selected as few-shot exemplars. The parameter $\lambda$ serves as an explicit diversity control factor, regulating the trade-off between semantic relevance to the query and diversity among the selected demonstrations.
A larger $\lambda$ emphasizes query relevance, while a smaller $\lambda$ increases the diversity characteristic, encouraging the selection of mutually distinctive cases.
\begin{align} 
\mathcal{S}_{\mathrm{case}}&=  \textrm{DiverseSim}(l^{query}, D^{domain}, n, \lambda)
\end{align}
Through this process, we obtain the top-$n$ semantically similar legal cases, forming the case-based exemplars for subsequent hybrid few-shot in-context learning.


\subsubsection{Entity-agnostic Few-shot Selection}

While semantic case-based retrieval provides relevant contextual demonstrations, legal texts are sometimes dominated by lengthy and highly specific entity mentions, such as detailed descriptions of illegal activities, harms, assets, and contract identifiers.
These entity-heavy expressions may distort similarity measurements and bias retrieval toward surface-level entity overlap rather than legally meaningful reasoning patterns. 

To mitigate this issue, we propose an entity-agnostic few-shot selection strategy based on template content. 
Specifically, given a legal case $l_i$ and a query case $l^{\mathrm{query}}$, we first apply a template function $template(\cdot)$ to transform both cases into entity-agnostic templates. This function substitutes concrete entity instantiations with their entity type, preserving the underlying legal relations and structure of the case.
The top-$m$ most relevant exemplars are retrieved by the \textit{DiverseSim} algorithm. This template-based retrieval mechanism enables the selection of structurally aligned demonstrations even when the original cases differ substantially in their surface entities.
The template-based similar exemplars are retrieved as:
\begin{align}
\mathcal{T}&= \{(\mathrm{template}(l_i), f_i) \}_{l_i \in \mathcal{D}^{domain}} \\
\mathcal{S}_{\mathrm{template}}&=  \textrm{DiverseSim}(l^{query}, \mathcal{T}, m, \lambda)
\end{align}
where $l_i$ and $f_i$ represent the legal case and the corresponding fact formulas, respectively. 
As in the case-based retrieval, the parameter $\lambda$ serves as a diversity score that balances relevance and diversity during exemplar selection.
This process yields the top-$m$ template-based exemplars, serving as entity-agnostic few-shot demonstrations.

\subsubsection{Diversity-aware Few-shot Combination}
We integrate the retrieved template-based exemplars (the top-$m$ templates) with the $n$ case-based exemplars into a unified few-shot prompt, forming a diversity-aware hybrid in-context learning environment. 
This complementary design enables the model to simultaneously leverage realistic legal contexts and structurally rich reasoning patterns, thereby improving robustness and generalization in legal logical inference.
\begin{align}
\mathcal{S}_{\mathrm{few}} &= \mathcal{S}_{\mathrm{case}} \cup \mathcal{S}_{\mathrm{template}} 
\end{align}
where the hyperparameters $n$ and $m$ control the balance between semantic and structural relevance in the in-context demonstrations, respectively. 

\subsection{Prompting Construction and Inference \label{few_shot}}

This subsection describes the construction of the input prompt using the selected exemplars from the previous step. As illustrated in Table~\ref{tab:prompt_hybrid}, the prompt is composed of two complementary types of demonstrations introduced in Section~\ref{sec:demon_ranking}: case-based exemplars and template-based exemplars. These demonstrations are concatenated sequentially to form a unified in-context learning prompt, which is then provided to the large language model for legal semantic parsing.
\begin{align}
f^{prediction} = \textrm{LLM-decode}(\textrm{prompting}(l^{query}, \mathcal{S}_{few}))
\end{align}
We defer a detailed analysis of the characteristics and effects of different few-shot retrieval strategies to the experimental section, where we systematically examine how various combinations of case-based and template-based exemplars influence model performance.

\begin{table}[htbp]
    \small
    \centering
    \caption{Prompting template combining case-based and template-based demonstrations. The \textcolor{blue}{blue} text indicates content that is replaced with data from the selected few-shot set, $\mathcal{S}_{\text{few}}$, while the \textcolor{red}{red} text denotes the content to be generated by the LLM. \label{tab:prompt_hybrid}}
    \begin{tabular}{|p{0.95\columnwidth}|}
        \hline
        \#\#\# You are an expert in the Semantic parsing task, which maps from legal cases to logical formulas (Note: following the exact function name defined in the fewshot samples).\\
        \\
        \#\#\# Input: \textcolor{blue}{{\{\{ Selected legal case, $l_i \in \mathcal{S}_{few}$ \}\}}} \\
        \#\#\# Logical Formulas Template: \textcolor{blue}{{\{\{ $\textrm{template}(f_i)$ where $ f_i \in \mathcal{S}_{few}$ \}\}}}  \\
        \#\#\# Output: \textcolor{blue}{{\{\{ Logical facts, $f_i \in \mathcal{S}_{few}$ \}\}}}  \\[0.2em]
        ... \\[0.2em]
        \#\#\# Input: \textcolor{blue}{{\{\{ Query legal case, $l^{query}$ \}\}}} \\
         \#\#\# Logical Formulas Template: \textcolor{red}{\{\{ $\textrm{template}(f^{query})$ \}\} } \\
         \textcolor{red}{\#\#\# Output: \{\{Logical facts,  $f^{query}$ \}\} }\\
        \hline
    \end{tabular}
\end{table}

\section{Experiments}
\subsection{Datasets and Experimental Settings}

To evaluate the generalization ability of the proposed framework, \textit{Legal2LogicICL}, we primarily conduct experiment on \textit{Legal2Proleg} dataset, which is an extended version of the \textit{Legal2ProlegV0} dataset. In addition, we also conduct extensive experiments on the LegalCaseNER dataset to highlight the differing characteristics between the two datasets. The detailed comparison among three datasets is presented in Table \ref{tab:legal2logicicl_dataset}. 

We evaluate our approach on a diverse set of representative large language models, including the open-source models \texttt{Qwen3-8B}, \texttt{Qwen3-14B} \cite{yang2025qwen3}, \texttt{Llama-3.1-8B-Instruct} \citep{grattafiori2024llama3herdmodels}, and \texttt{Phi-4} \cite{abdin2024phi}, as well as the proprietary model \texttt{gpt-5.2} accessed via the OpenAI API. 
In our implementation of \textit{DiverSim}, \texttt{Qwen/Qwen3-Embedding-8B} model was used to encode the query and compute the similarity. 
All experiments are conducted with \textbf{five} different random seeds, and results are reported as the average performance across runs to reduce variance due to stochastic sampling.


\subsection{Evaluation Metric \label{sec:eval_metric}}

For evaluation, we adopt exact match accuracy as the primary evaluation metric following previous works \citep{zin2023improving,phuogdata} and Semantic-aware Evaluation.
This choice is motivated by the strict syntactic and compositional constraints of PROLEG fact formulas, where even minor deviations render a formula invalid for legal reasoning. Accordingly, a prediction is counted as correct only if the entire generated formula--including all PROLEG functions and entities--exactly matches the reference:
\begin{align}
    Accuracy  = \frac{1}{N} \sum_{i=1}^{N} \mathbb{I}\!\left(f_i^{\text{pred}} = f_i^{\text{gold}}\right)
\end{align}
where $N$ is the number of evaluation samples. It is worth noting that \textit{LegalCaseNER} \citep{zin2023improving} formulates the problem as an NER task, where correctness is defined by the accurate identification of all target entities. In contrast, our method directly translates a legal case description into a complete PROLEG fact formula, which necessitates a stricter, sentence-level exact match evaluation to faithfully reflect the precision requirements of automated legal reasoning.

In addition, to provide a more comprehensive evaluation beyond exact matching, we introduce a semantic-aware metric, Soft-Match Accuracy, that preserves structural correctness while relaxing surface-form constraints on entities. Specifically, we require an exact match on the logical structure of the predicted formula, and measure semantic similarity between predicted $e_k^{pred}$ and gold entities $e_k^{gold}$ using cosine similarity:
\begin{align}
Soft\textrm{-}Match\,\,Acc. &=  \frac{1}{N} \sum_{i=1}^{N} \Big( 
\mathbb{I}\big(\mathrm{struct}(f_i^{\text{pred}})=\mathrm{struct}(f_i^{\text{gold}})\big) \quad\backslash \notag \\
&\quad\,\,\,\times \frac{1}{K} \sum_{k=1}^{K}
\textrm{sim}\big(\mathrm{Emb}(e_{i,k}^{\text{pred}}),\mathrm{Emb}(e_{i,k}^{\text{gold}})\big)\Big) 
\end{align}
where $K$ denotes entity index in legal fact formulas, $\text{struct}(\cdot)$ denotes the structural form of logical expressions with entity information removed, and $\text{Emb}(\cdot)$ represents the embedding function using \texttt{Qwen/Qwen3-Embedding-8B} model. This metric enables a more fine-grained evaluation by accounting for semantic equivalence between entities, while strictly enforcing structural correctness.

\subsection{Experimental Results}





\subsubsection{Performance Comparison under Varying Training Data Proportions}

\begin{figure}[htbp]
    \centering
    \begin{tikzpicture}
        \begin{axis}[
            width=0.46\textwidth,
            height=0.28\textwidth,
            axis lines=middle,
            xmin=0.1, xmax=1.07,
            ymin=0, ymax=1.07,
            xlabel={\#seen data},
            ylabel={Acc.},
            ylabel style={yshift=6pt},
            xlabel style={yshift=-14pt,xshift=10pt},
            xtick={0.2,0.4,0.5,0.6,0.8},
            legend style={at={(0.79,0.7)}, anchor=north, align=left, legend columns=1},
            legend cell align={left},
            every axis plot/.append style={thick}
        ]

        \addplot[
            opacity=0.75,
            mark=diamond*,
            color=orange,
             very thick, nodes near coords, 
        ] coordinates {
            (0.2, 0.8326633166)
            (0.4, 0.9172529313)
            (0.5, 0.9259557344)
            (0.6, 0.9412060302)
            (0.8, 0.9497487437)
        };
        \addlegendentry{\textit{(ICL)} \texttt{Phi-4 14B} (ours)} 
        \addplot[color=orange,opacity=0.25,mark=none,forget plot] coordinates {(0.8, 0.9246231156) (0.6, 0.9170854271) (0.5, 0.9134808853) (0.4, 0.8994974874) (0.2, 0.8366834171)};
        \addplot[color=orange,opacity=0.25,mark=none,forget plot] coordinates {(0.8, 0.9346733668) (0.6, 0.9447236181) (0.5, 0.9275653924) (0.4, 0.9145728643) (0.2, 0.8454773869)};
        \addplot[color=orange,opacity=0.25,mark=none,forget plot] coordinates {(0.8, 0.9648241206) (0.6, 0.959798995) (0.5, 0.9436619718) (0.4, 0.9195979899) (0.2, 0.8278894472)};
        \addplot[color=orange,opacity=0.25,mark=none,forget plot] coordinates {(0.8, 0.959798995) (0.6, 0.9422110553) (0.5, 0.9215291751) (0.4, 0.9229480737) (0.2, 0.8165829146)};
        \addplot[color=orange,opacity=0.25,mark=none,forget plot] coordinates {(0.8, 0.9648241206) (0.6, 0.9422110553) (0.5, 0.9235412475) (0.4, 0.9296482412) (0.2, 0.8366834171)};
        
        \addplot[
            opacity=0.75,
            mark=triangle*,
            color=blue,
              thick,  
        ] coordinates {
            (0.2, 0.826325)
            (0.4, 0.89914)
            (0.5, 0.9047333333)
            (0.6, 0.9216)
            (0.8, 0.93164)
        };
        \addlegendentry{\textit{(ICL)} \texttt{Qwen3-14B}  (ours)}
        \addplot[color=blue,opacity=0.25,mark=none,forget plot] coordinates {(0.8, 0.9095) (0.6, 0.9045) (0.5, 0.9054) (0.4, 0.8911) (0.2, 0.8317)};
        \addplot[color=blue,opacity=0.25,mark=none,forget plot] coordinates {(0.8, 0.9447) (0.6, 0.9246) (0.5, 0.9014) (0.4, 0.9028) (0.2, 0.8417)};
        \addplot[color=blue,opacity=0.25,mark=none,forget plot] coordinates {(0.8, 0.9397) (0.6, 0.9397) (0.5, 0.9135) (0.4, 0.9045) (0.2, 0.8291)};
        \addplot[color=blue,opacity=0.25,mark=none,forget plot] coordinates {(0.8, 0.9296) (0.6, 0.9121) (0.5, 0.8893) (0.4, 0.8978) (0.2, 0.8028)};
        \addplot[color=blue,opacity=0.25,mark=none,forget plot] coordinates {(0.8, 0.9347) (0.6, 0.9271) (0.5, 0.9075) (0.4, 0.8995) (0.2, 0.8241)};

        \addplot[densely dashed, cyan, domain=0.2:0.8] {0.8809};
        \addlegendentry{\textit{(ICL)} \texttt{gpt-5.2} (ours)}

        \addplot[
            opacity=0.75,
            mark=triangle*,
            color=red,
             thick
        ] coordinates {
            (0.2, 0.7379)
            (0.4, 0.8224)
            (0.5, 0.8322)
            (0.6, 0.8472)
            (0.8, 0.8663)
        };
        \addlegendentry{\textit{(ICL)} \texttt{Qwen3-8B} (ours) }
        \addplot[color=red,opacity=0.25,mark=none,forget plot] coordinates {(0.8, 0.8693) (0.6, 0.8492) (0.5, 0.8571) (0.4, 0.8727) (0.2, 0.7575)};
        \addplot[color=red,opacity=0.25,mark=none,forget plot] coordinates {(0.8, 0.8844) (0.6, 0.8391) (0.5, 0.8249) (0.4, 0.8157) (0.2, 0.7475)};
        \addplot[color=red,opacity=0.25, mark=none,forget plot] coordinates {(0.8, 0.8643) (0.6, 0.8693) (0.5, 0.8249) (0.4, 0.8141) (0.2, 0.7198)};
        \addplot[color=red,opacity=0.25, mark=none,forget plot] coordinates {(0.8, 0.8693) (0.6, 0.8191) (0.5, 0.8209) (0.4, 0.799) (0.2, 0.7186)};
        \addplot[color=red, opacity=0.25,mark=none,forget plot] coordinates {(0.8, 0.8442) (0.6, 0.8592) (0.5, 0.833) (0.4, 0.8107) (0.2, 0.7462)};

        \addplot[
            opacity=0.75,
            mark=diamond*,
            color=black
        ] coordinates {
            (0.2, 0.1661195957)
            (0.4, 0.6369082973)
            (0.5, 0.7658362295)
            (0.6, 0.8258732407)
            (0.8, 0.8851420101)
        };
        \addlegendentry{\textit{(SFT)} LegalCaseNER \citep{zin2023improving} }
        \addplot[ color=black,opacity=0.25,mark=none,forget plot] coordinates {(0.8, 0.8442211055) (0.6, 0.7713567839) (0.5, 0.6921529175) (0.4, 0.634840871) (0.2, 0.1496855346)};
        \addplot[ color=black,opacity=0.25,mark=none,forget plot] coordinates {(0.8, 0.9137055838) (0.6, 0.8181818182) (0.5, 0.7797979798) (0.4, 0.6336134454) (0.2, 0.1551071879)};
        \addplot[ color=black,opacity=0.25,mark=none,forget plot] coordinates {(0.8, 0.8994974874) (0.6, 0.8316582915) (0.5, 0.7927565392) (0.4, 0.6649916248) (0.2, 0.1746231156)};
        \addplot[ color=black,opacity=0.25,mark=none,forget plot] coordinates {(0.8, 0.8888888889) (0.6, 0.8513853904) (0.5, 0.7717171717) (0.4, 0.6346801347) (0.2, 0.1803278689)};
        \addplot[ color=black,opacity=0.25,mark=none,forget plot] coordinates {(0.8, 0.8793969849) (0.6, 0.8567839196) (0.5, 0.7927565392) (0.4, 0.6164154104) (0.2, 0.1708542714)};
        
        \end{axis}
    \end{tikzpicture}
    \caption{Performance comparison between NER-based method \textit{LegalCaseNER} and few-shot learning methods \textit{Legal2LogicICL} on the Legal2Proleg dataset with respect to the portion of seen data. Results are averaged over five random seeds, corresponding to the shaded lines.\label{fig:leng_analysis}}
\end{figure}
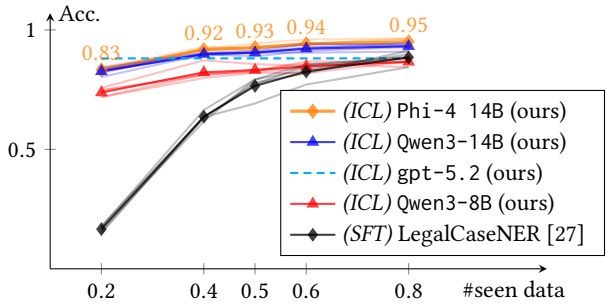

We conduct a comparative evaluation between NER-based supervised models and our \textit{Legal2LogicICL} framework on the \textbf{Legal2Proleg} dataset under varying training data splits, with the proportion of seen data ranging from 0.2 to 0.8. To construct these splits (seen data), we apply a repeated hold-out strategy. In the dataset construction stage, each template is used to generate one unique sample, ensuring that different samples correspond to different templates. During data splitting, the dataset is randomly partitioned into training and test sets while ensuring that templates appearing in the test set do not occur in the training set, thereby preventing template overlap across splits. 
As illustrated in Figure~\ref{fig:leng_analysis}, when the seen data ratio is relatively high (0.8), the NER-based model \textit{LegalCaseNER} achieves a competitive average performance of 88.51\%, compared to 95\% obtained by our method (phi-4). However, as the amount of labeled training data decreases, the performance gap between the two approaches becomes increasingly pronounced. In the low-resource setting with a seen data ratio of 0.2, the accuracy of the NER-based model drops sharply to 16.61\%, whereas our \textit{Legal2LogicICL} maintains a substantially higher performance of 83\%.
These results demonstrate that our \textit{Legal2LogicICL} framework exhibits strong robustness across different data availability settings and is significantly less sensitive to the proportion of labeled training data. This property is particularly critical for real-world deployment, where user inputs are inherently open-ended and cannot be exhaustively covered by pre-annotated training cases. 
In contrast, the \textit{LegalCaseNER} approach degrades severely in the presence of a large amount of unseen data, indicating its limited generalization capability under low-resource conditions. This limitation poses a significant risk in practical applications, as it is unrealistic to guarantee sufficient labeled training data for every possible legal scenario encountered in free-form user inputs.

For the ChatGPT experiment (\texttt{gpt-5.2}), due to the high cost of OpenAI services, we conducted evaluations only under the seen-data ratio of 0.6 (dashed line). The results demonstrate that our \textit{Legal2LogicICL} framework has the potential to work effectively across different LLMs. Further analysis of the \texttt{gpt-5.2} results is provided in Section~\ref{sec:error_analysis_gpt52}, with representative error examples presented in Table~\ref{tab:legal2proleg_error_example}.

\subsubsection{Effect of Diversity Control Parameter ($\lambda$)}
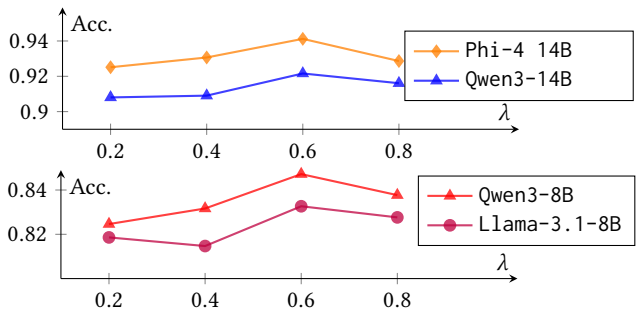
\begin{figure}[htbp]
    \centering
    \begin{tikzpicture}
        \begin{axis}[
            width=0.43\textwidth,
            height=0.18\textwidth,
            axis lines=middle,
            xmin=0.1, xmax=1.05,
            ymin=.89, ymax=.959,
            xlabel={$\lambda$},
            ylabel={Acc.},
            xtick={0.2,0.4,0.6,0.8},
            legend style={at={(1.,0.81)}, anchor=north, align=left, legend columns=1},
            legend cell align={left},
            every axis plot/.append style={thick}
        ]


        \addplot[
            opacity=0.75,
            mark=diamond*,
            color=orange,
             thick
        ] coordinates {
            (0.2, 0.9251256281)
            (0.4, 0.9306532663)
            (0.6, 0.9412060302)
            (0.8, 0.9286432161)
            
        };
        \addlegendentry{\texttt{Phi-4 14B \quad\,\,\,}}

        \addplot[
            opacity=0.75,
            mark=triangle*,
            color=blue,
             thick
        ] coordinates {
            (0.2, 0.908040201)
            (0.4, 0.9090452261)
            (0.6, 0.9216)
            (0.8, 0.916080402)
        };
        \addlegendentry{\texttt{Qwen3-14B}}


        \end{axis}
    \end{tikzpicture}
    \begin{tikzpicture}
        \begin{axis}[
            width=0.43\textwidth,
            height=0.17\textwidth,
            axis lines=middle,
            xmin=0.1, xmax=1.05,
            ymin=.80, ymax=.849,
            xlabel={$\lambda$},
            ylabel={Acc.},
            xtick={0.2,0.4,0.6,0.8},
            legend style={at={(1.02,0.95)}, anchor=north, align=left, legend columns=1},
            legend cell align={left},
            every axis plot/.append style={thick}
        ]

        \addplot[
            opacity=0.75,
            mark=triangle*,
            color=red,
             thick
        ] coordinates {
            (0.2, 0.8246231156)
            (0.4, 0.8316582915)
            (0.6, 0.84718	)
            (0.8, 0.8376884422)
        };
        \addlegendentry{\texttt{Qwen3-8B}}

        \addplot[
            opacity=0.75,
            mark=*,
            color=purple,
             thick
        ] coordinates {
            (0.2, 0.8185929648)
            (0.4, 0.8146984925)
            (0.6, 0.8326633166)
            (0.8, 0.827638191)
        };
        \addlegendentry{\texttt{Llama-3.1-8B}}

        \end{axis}
    \end{tikzpicture}
    \caption{Performance comparison on the effect of the diversity control parameter ($\lambda$) across diverse LLMs under identical ICL experimental settings (seen data rate $= 0.6$). Results are averaged over five random seeds for each setting.  }
    \label{fig:lambda_analysis}
\end{figure}

As shown in Figure~\ref{fig:lambda_analysis},  a moderate setting of the diversity control parameter ($\lambda = 0.6$) consistently achieves the highest diversity across all evaluated model backbones.
This observation highlights the importance of maintaining an appropriate level of diversity in latent semantic vector representations (deep-level). The results indicate that all LLMs exhibit the same performance trend. When the $\lambda$ value is decreased (e.g., $\lambda = 0.2$), diversity is emphasized, yielding exemplars that remain relevant to the query while being more diverse with respect to one another. Conversely, when the $\lambda$ value is increased (e.g., $\lambda = 0.8$), similarity is prioritized, resulting in exemplars that are more similar to the query and to each other.


\subsubsection{Effect of hybrid few-shot strategy}
As shown in Table~\ref{tab:kshot_ablation}, under identical data rate and diversity control settings, the choice of $k$-shot strategy has a substantial impact on performance. Specifically, \textit{3c} and \textit{5c} denote three and five case-based few-shot examples, respectively, while \textit{3c+3t} ($n=3, m=3$) represents our proposed hybrid strategy combining three case-based and two template-based examples.

Here, in the setting of 5c, comparison of our DiverseSim ($\lambda = 0.6$) with a cosine similarity–based method ($\lambda = 1$, where diversity is disabled and only similarity-based examples are selected) shows that performance consistently drops across LLMs. This confirms that diversity-aware retrieval leads to better performance than purely similarity-based selection, demonstrating the effectiveness of the diversity characteristic.

Across all evaluated configurations, the proposed \textit{3c+3t} strategy consistently achieves the best performance across four backbone models.
These results indicate that increasing the number of semantically retrieved examples improves performance, but the model remains sensitive to entity-induced retrieval bias. The hybrid few-shot strategy incorporating both semantic cases and template-based entity-agnostic exemplars consistently yields further gains, even under the same total number of few-shot examples (k=5). This confirms that the proposed hybrid retrieval strategy effectively balances semantic relevance and structural diversity.
\begin{table}[htbp]
\caption{Ablation study on the impact of different $k$-shot strategies under identical seen data rate ($= 0.6$).}
\label{tab:kshot_ablation}
\centering
\small
\resizebox{\linewidth}{!}{
    \begin{tabular}{l c c c c c c}
    \toprule
    \textbf{$k$-shot} &  $\lambda$ & \texttt{Qwen3-8B}  & \texttt{Llama-3.1-8B}  & \texttt{Qwen3-14B}   & \texttt{Phi-4} \\
    \midrule
    3c  & 0.6    &   0.7075 & 0.7824 &  0.8437 & 0.8894  \\
    5c & 1.0     &  0.7899 & 0.8171 & 0.8930 &  0.9141  \\  
    5c   &  0.6  &  0.8005 & 0.8246 & 0.8955 &  0.9216  \\ 
    3c+3t&  0.6   & \textbf{0.8472} & \textbf{0.8327} &  \textbf{0.9216} &  \textbf{0.9412} \\ 
    \bottomrule
    \end{tabular}
}
\end{table}

\subsubsection{Effect of Entity Name Bias} 
As shown in Table~\ref{tab:exp_ner_data}, we evaluate our proposed \textit{Legal2LogicICL} framework on the LegalCaseNER dataset \citep{zin2023improving} and re-implement their proposed approach.
The SFT-based NER model achieves near-perfect performance on this dataset. This outcome can be attributed to two primary factors: (1) the dataset contains a small number of patterns, with the smallest template vocabulary but the largest number of samples among the evaluated datasets (Table~\ref{tab:legal2logicicl_dataset}); and (2) the training and test sets share the same templates, which makes the model prone to overfitting and limits its robustness in real-world scenarios. 
In contrast, the main source of errors in our ICL-based approach  \textit{Legal2LogicICL} stems from minor variations in entity surface forms, such as missing articles (e.g., “a” or “the”). These errors are closely related to annotation consistency in the dataset and do not affect the logical soundness of the resulting legal reasoning representations. 
\begin{table}[htbp]
\caption{Impact of entity name bias on model performance.}
\label{tab:exp_ner_data}
\centering
\resizebox{\linewidth}{!}{
    \begin{tabular}{p{0.17\linewidth} c c c c c c}
    \toprule
    \textbf{Seen data} &   \textit{Legal2LogicICL}   &  \textit{Legal2LogicICL}  &  \textit{(SFT) LegalCaseNER}  \\
     &   (\texttt{Qwen3-8B}) &  (\texttt{Qwen3-14B})  & (\texttt{Roberta-large})    \\
    \midrule
    40\%      &   0.7973 & 0.9000 &  \textbf{0.9957}\\
    60\%      &   0.8353 & 0.8975 & \textbf{0.9988}  \\
    80\%      &   0.8466 & 0.8981 &  \textbf{0.9953 } \\
    \bottomrule
    \end{tabular}
}
\end{table}

\subsection{Result Analysis}

\subsubsection{Diversity Analysis in Latent Semantic Vector Space} 
\begin{figure}[htbp]
    \centering
    \includegraphics[width=0.99\linewidth, keepaspectratio, 
            trim={1.1cm 0.6cm 0.8cm  0.7cm}, page=1, clip=true]{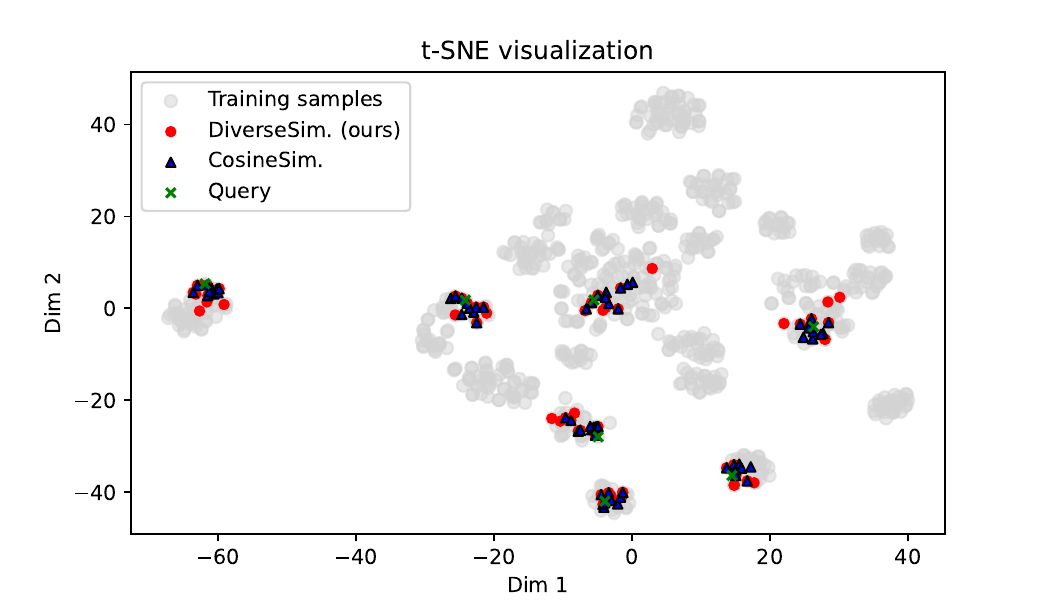} 
    \caption{Visualization of our ranking method (DiverseSim) with $\lambda=0.6$. Each data point represents a sample in the Legal2Proleg dataset. The highlighted points indicate the few-shot exemplars selected by DiverseSim and by cosine similarity with respect to the query, respectively.}
    \label{fig_viz}
\end{figure}

Here, we analyze the effect of \textit{DiverseSim} in comparison with conventional cosine similarity for the few-shot exemplar selection process. Figure~\ref{fig_viz} visualizes the semantic representations of all legal cases in the training set (seen data) using t-SNE dimensionality reduction \citep{JMLR:v9:vandermaaten08a}. Different clusters correspond to distinct legal issues or contract types. This visualization demonstrates that the exemplars selected by DiverseSim exhibit are more widely distributed within each semantic cluster, demonstrating greater semantic diversity, whereas cosine similarity-based retrieval tends to concentrate exemplars within a narrower region around the query representation.

\subsubsection{ Diversity Analysis of Retrieved Few-shot Exemplars}

To better understand the effectiveness of the proposed hybrid few-shot strategy, we analyze exemplar diversity under two retrieval settings: (i) semantic retrieval based on cosine similarity, and (ii) template-aware, entity-agnostic selection based on template-level matching. Table~\ref{tab:diversity_case_study} presents representative examples from both methods.
As illustrated in the table, semantic retrieval tends to select exemplars that are highly similar to the query at the surface level, including identical entities (e.g., the same names such as Alex and Jordan) and evaluative expressions (e.g., ``What an inventive version!''), as well as nearly identical narrative structures. Beyond entity overlap, most retrieved examples exhibit minimal structural variation, indicating that the retrieval process is dominated by entity-level similarity and provides limited diversity in legal reasoning patterns.

In contrast, the proposed template-aware entity-agnostic retrieval strategy abstracts away from concrete entities and emphasizes structural variations across cases. While the retrieved exemplars share the same high-level legal structure as the query case, they exhibit greater diversity in reasoning paths and predicate instantiations. This qualitative difference indicates that template-level matching effectively increases structural diversity within the few-shot context, complementing semantic relevance and contributing to more robust in-context legal reasoning.

\begin{table*}[htbp]
\caption{Qualitative comparison of few-shot exemplars retrieved by semantic-only and template-aware strategies.}
\centering
\small
\resizebox{\linewidth}{!}{
\begin{tabular}{l p{0.5\linewidth} p{0.54\linewidth}}
\toprule
\cellcolor{gray!15}\textbf{Case} & 
\multicolumn{2}{c}{\cellcolor{gray!15}\textbf{Content}}  \\
\midrule
\textbf{Query Case} &
\multicolumn{2}{p{1.05\linewidth}}{
Artist Jordan admires musician Alex and habitually listened to Alex’s composition. Jordan released a new composition by adjusting the vocals while maintaining the hook, sharing it as his own on SoundCloud. Eventually, Alex acknowledged, ``What an inventive version!'', causing the composition to exceed one million plays.
} \\
\midrule
\cellcolor{gray!15}\textbf{Exemplar} &
\cellcolor{gray!15}\textbf{Semantic-based Retrieved Exemplars} &
\cellcolor{gray!15}\textbf{Template-based Retrieved Exemplars} \\
\midrule
Example\_1
&
Inspired by Alex, Jordan who habitually listened to Alex’s composition, created a new melody by adjusting the vocals while maintaining the hook. Posting it on his Facebook page, Alex acknowledged, ``What an inventive version!'', it soon attracted a million likes.
&
Musician V is admired by person U and often vibed to V’s rhythm. U crafted a fresh tune by preserving the tempo but altering the arrangement, subsequently sharing it on Instagram as U's composition. Eventually, V said, ``This is captivating,'', leading the track to amass over two million streams.
\\
\hline
Example\_2
&
Musician Casey has a fan named Chris who played regularly Casey’s tune. Chris created a new track by changing the melody slightly but keeping the hook and uploaded it as Chris's own on Instagram. Subsequently, Casey responded, ``Amazing interpretation!'', leading to the track amassing a million streams.
&
T is a fan of musician U and heavily sampled U’s piece. T crafted a new tune while keeping the vocal style retained but changed the tempo. T shared this on his SoundCloud as original.
\\
\hline
Example\_3
&
Artist Jamie has a follower in Sam who practiced daily Jamie’s single. Sam composed a new melody by modifying the chords but keeping the pre-chorus and published it on TikTok as Sam's piece. Finally, Jamie expressed, ``Fantastic transformation!'', resulting in the track garnering over three million plays.
&
Inspired by Alex, Jordan who habitually listened to Alex’s composition, created a new melody by adjusting the vocals while maintaining the hook. Posting it on his Facebook page, Alex acknowledged, ``What an inventive version!'', it soon attracted a million likes.
\\
\bottomrule
\end{tabular}
}
\label{tab:diversity_case_study}
\end{table*}






\subsubsection{Error Analysis of LegalCaseNER}
\label{sec:error_legalcasener}

To better understand the limitations of NER-based legal semantic parsing methods, we conduct a qualitative error analysis of \textit{LegalCaseNER}.
Although NER-based approaches are capable of identifying surface-level entities in legal texts, we observe that they often fail to capture legally meaningful entities and relations, particularly when training data is limited.
Table~\ref{tab:ner_error_example} presents a representative example illustrating several typical failure modes. For clarity, we highlight incorrect entity predictions in red and their corresponding gold-standard entities in blue, while leaving correctly predicted entities unmarked.

\paragraph{Shallow and Incorrect Entity Boundary Detection.}
As shown in the example, the gold entity \textit{(the conference room)[Object]} is incorrectly predicted as \textit{(the)[Asset] conference room}.
Here, the model assigns the entity label \textit{Asset} to a function word (\textit{the}), while excluding the actual legal object (\textit{conference room}) from the entity span.
This behavior indicates that the NER model relies heavily on local lexical cues rather than semantically coherent spans, leading to fundamentally incorrect entity grounding.
Such boundary errors directly propagate to downstream logical parsing, resulting in invalid or incomplete fact formulas.

\paragraph{Failure under Overlapping or Semantically Composite Entities.}
Legal texts frequently express harms, obligations, or violations as semantically rich clauses rather than isolated noun phrases.
In the gold annotation, the harm is represented as a single composite entity:
\textit{(Lucas’s reputation was compromised as clients began avoiding the venue due to the unexpected activities)[Harm]}.
However, LegalCaseNER fragments this span into multiple unrelated entities, such as \textit{(Lucas’s)[Lender]} and \textit{(the venue)[Object]}.
This demonstrates that NER-based models struggle when a legally salient concept subsumes multiple surface-level entities, making it difficult to recover higher-order legal facts such as reputation damage.

\paragraph{Confusion between Semantically Similar Entity Types.}
We further observe frequent confusion between temporally related entity types.
In the example, the discovery time \textit{(2023/06/30)[T\_discovery]} is incorrectly labeled as \textit{(2023/06/30)[T\_due]}.
This error reflects an inherent limitation of flat NER labeling schemes in distinguishing legally distinct but lexically similar temporal concepts.
Such confusion can critically affect legal reasoning, as different temporal roles often trigger different legal consequences.

\paragraph{Discussion.}
These reveal fundamental limitations of NER-based approaches.
By design, LegalCaseNER focuses on identifying surface-level spans and assigning fixed entity labels, without explicitly modeling underlying legal structures or reasoning patterns.
As a result, it is particularly fragile when faced with complex legal narratives, overlapping entity semantics, and limited training data.
In contrast, our proposed framework avoids explicit entity labeling during inference and instead leverages structurally grounded, template-level reasoning through in-context learning, making it more robust to such entity-induced noise.

\begin{table}[htbp]
\caption{Representative Errors of LegalCaseNER (incorrect predictions in red, corresponding gold entities in blue)}
\label{tab:ner_error_example}
\centering
\small
\resizebox{\linewidth}{!}{
\begin{tabular}{  p{1\linewidth}}
\toprule
\cellcolor{gray!15}\textbf{(Prediction)} \\
(Lucas)[Lender] provided 
\textcolor{red}{(the)[Asset]} conference room 
to (Mia)[Borrower] under (conference-rental8)[Agreement] for enhancing their product offerings. 
However, (Mia)[Borrower] did not utilize ...,
\textcolor{red}{(Lucas’s)[Lender]} reputation was compromised as clients began avoiding 
\textcolor{red}{(the venue)[Object]} due to the unexpected activities. 
The actions of (Mia)[Borrower] came to light on 
\textcolor{red}{(2023/06/30)[T\_due]}. \\
\midrule
\cellcolor{gray!15}\textbf{(Gold Ouput)}  \\
(Lucas)[Lender] provided 
\textcolor{blue}{(the conference room)[Object]} 
to (Mia)[Borrower] under (conference-rental8)[Agreement] for enhancing their product offerings. 
However, (Mia)[Borrower] did not utilize ..., 
\textcolor{blue}{(Lucas’s reputation was compromised as clients began avoiding the venue due to the unexpected activities)[Harm]}. 
The actions of (Mia)[Borrower] came to light on 
\textcolor{blue}{(2023/06/30)[T\_discovery]}. \\
\bottomrule
\end{tabular}
}
\end{table}

\subsubsection{Error Analysis of Legal2LogicICL}\label{sec:error_analysis_gpt52}

\begin{table}[htbp]
\caption{Representative Errors of Legal2LogicICL (incorrect entities in red, corresponding gold entities in blue)}
\label{tab:legal2proleg_error_example}
\centering
\small
\resizebox{\linewidth}{!}{
\begin{tabular}{ p{1.2\linewidth}}
\toprule
\multicolumn{1}{r}{\cellcolor{gray!15}\textbf{(Prediction)}}   \\ 
\texttt{borrower('Mason').} \newline
\texttt{ lender('Emma').} \newline
\texttt{ owned\_by(\textcolor{red}{'the laptop'}, 'Emma').} \newline
\texttt{ borrowing\_agreement('lease78').} \newline
\texttt{ damage\_fact('Mason', \textcolor{red}{'the laptop'}).} \newline
\texttt{ repair\_payment\_request\_fact('Mason','Emma', \textcolor{red}{'the laptop'}).} \newline
\texttt{ repair\_request\_fact('Mason','Emma', \textcolor{red}{'the laptop'}, '2024/01/10').} \newline
demob :- block(right\_to\_dispute\_repair\_demand('Emma','Mason', \textcolor{red}{'the laptop'}, 'lease78')). \\ \midrule
\multicolumn{1}{r}{\cellcolor{gray!15}\textbf{(Gold)}}   \\  
\texttt{borrower('Mason').} \newline
\texttt{ lender('Emma').} \newline
\texttt{ owned\_by(\textcolor{blue}{'a laptop'}, 'Emma').} \newline
\texttt{ borrowing\_agreement('lease78').} \newline
\texttt{ damage\_fact('Mason', \textcolor{blue}{'a laptop'}).} \newline
\texttt{ repair\_payment\_request\_fact('Mason','Emma', \textcolor{blue}{'a laptop'}).} \newline
\texttt{ repair\_request\_fact('Mason','Emma', \textcolor{blue}{'a laptop'}, '2024/01/10').} \newline
demob :- block(right\_to\_dispute\_repair\_demand('Emma','Mason', \textcolor{blue}{'a laptop'}, 'lease78')). \\
\bottomrule
\end{tabular}
}
\end{table}
In this section, we conduct an error analysis of the outputs generated by ChatGPT-5 (\texttt{gpt-5.2}).
We manually inspect the predictions produced under the setting of $\lambda = 0.6$ with a seen-data rate of 60\%, using one representative random seed (overall result shown in Figure~\ref{fig:leng_analysis}, dashed line).
Our analysis reveals that a frequent class of errors corresponds to minor surface-form inconsistencies, as illustrated in Table~\ref{tab:legal2proleg_error_example}. These errors mainly involve confusion between indefinite and definite articles (e.g., “a” and “the”), missing or redundant articles, as well as occasional omissions of punctuation symbols such as periods. Such errors account for 17 out of 38 incorrect cases, corresponding to approximately 5\% of the total predictions.
Although these cases are counted as errors under our strict evaluation protocol, they are largely attributable to annotation inconsistencies in the dataset, which was labeled by multiple annotators. Importantly, these surface-level variations may not affect the underlying legal semantics or the correctness of logical reasoning, and would not lead to erroneous conclusions in practical legal reasoning systems. We therefore report these results to provide a transparent and comprehensive assessment of model behavior.

\paragraph{Effect of Miss Matching Entity Name.} In order to evaluate the affect of miss matching entity name in the semantic parsing process of our \textit{Legal2LogicICL} framework,  we report the different between two evaluation metrics Exact-Match and Soft-Match (described in sec \ref{sec:eval_metric}) in Table~\ref{tab:semanticeval}. Similar to the observation in \texttt{gpt-5.2}, we found that, our framework archived high performance in parsing the structure of logical expressions. Although the entity name some how hard to exactly match with the gold data but it can keep the major semantic meaning, which may not affect to the legal reasoning process, which this system can be contributed to. 

\begin{table}[htbp]
\caption{Comparison on two evaluation metrics, Exact-Match Acc. and Soft-Match Acc., under settings of  $3c+3t$ and $\lambda=0.6$.}
\label{tab:semanticeval}
\centering
\small
\resizebox{\linewidth}{!}{
    \begin{tabular}{l lllll}
    \toprule
    \textbf{Eval.Metric} &  \textbf{\texttt{Qwen3-8B}}  & \textbf{\texttt{Llama-3.1-8B}}  & \textbf{\texttt{Qwen3-14B}}   & \textbf{\texttt{Phi-4}} \\
    \midrule
    Exact-Match Acc.  & {84.72} & {83.27} &  {92.16} &  {94.12} \\ 
    Soft-Match Acc.   &  {92.18}$_{(+7.46)}$ & {97.68}$_{(+14.44)}$ &  {97.52}$_{(+4.91)}$ & {98.00}$_{(+3.88)}$  \\ 
    \bottomrule
    \end{tabular}
}
\end{table}

\section{Conclusion}
This paper presents \textit{Legal2LogicICL}, a novel in-context learning method for transforming natural language legal case into formal PROLEG logical facts. A key contribution of this work lies in the design of a diversity-aware hybrid few-shot retrieval strategy, implemented via the DiverseSim ranking algorithm, which jointly considers semantic case-level similarity and entity-agnostic template-level matching. By balancing contextual relevance with structural diversity, Legal2LogicICL constructs more informative and robust in-context demonstrations, leading to more accurate and stable logical rule generation.
In addition, we introduce \textit{Legal2Proleg}, a new benchmark dataset featuring annotated legal rules and corresponding PROLEG logical formulas, to support the systematic evaluation of legal semantic parsing. Experimental results across both open- and closed-source LLMs demonstrate that Legal2LogicICL consistently improves accuracy, stability, and generalization in parsing natural language legal cases into formal logical representations.
Overall, this work provides a practical and effective few-shot pa
radigm for explainable and reliable legal reasoning. Beyond the specifics of semantic parsing, the proposed retrieval strategy and in-context learning framework are generalizable and can be seamlessly integrated into legal reasoning systems, offering a scalable and interpretable foundation for future research in legal AI.

\section*{Acknowledgments}
This work was supported by the ``R\&D Hub Aimed at Ensuring Transparency and Reliability of Generative AI Models'' project of the MEXT, by JSPS KAKENHI Grant Numbers, 25H00522 and 25H01112,  and by JST as part of Adopting Sustainable Partnerships for Innovative Research Ecosystem (ASPIRE), Grant Number JPMJAP25B2. 
\bibliographystyle{ACM-Reference-Format}
\bibliography{ref}




\end{document}